# Pre-Selection of Independent Binary Features: An Application to Diagnosing Scrapie in Sheep

**L.I. Kuncheva**
School of Informatics,
University of Wales,
Bangor, Gwynedd,
LL57 1UT, UK
*l.i.kuncheva@bangor.ac.uk*

**C.J. Whitaker**
School of Informatics,
University of Wales,
Bangor, Gwynedd,
LL57 1UT, UK
*c.j.whitaker@bangor.ac.uk*

**P.D. Cockcroft**
Department of Clinical
Veterinary Medicine,
University of Cambridge,
CB3 0ES, UK
*pdc24@hermes.cam.ac.uk*

**Z.S.J. Hoare**
School of Informatics,
University of Wales,
Bangor, Gwynedd,
LL57 1UT, UK
*z.s.hoare@bangor.ac.uk*

## Abstract

Suppose that the only available information in a multi-class problem are expert estimates of the conditional probabilities of occurrence for a set of binary features. The aim is to select a subset of features to be measured in subsequent data collection experiments. In the lack of any information about the dependencies between the features, we assume that all features are conditionally independent and hence choose the Naive Bayes classifier as the optimal classifier for the problem. Even in this (seemingly trivial) case of complete knowledge of the distributions, choosing an optimal feature subset is not straightforward. We discuss the properties and implementation details of Sequential Forward Selection (SFS) as a feature selection procedure for the current problem. A sensitivity analysis was carried out to investigate whether the same features are selected when the probabilities vary around the estimated values. The procedure is illustrated with a set of probability estimates for Scrapie in sheep.

## 1 Introduction

We shall call "traditional" data sets those that observe values of a number of features ($n$) for each of a number of objects ($N$). There are problems in which a traditional data set is not easily available. Instead, there is expert knowledge about the problem in the form of a (possibly large) set of relevant features and hypothesised relationships of these features with the classes. Examples of such problems can be found in veterinary medicine. Expert knowledge about the relationships between clinical signs and diseases forms the basis for expert systems [17].

We consider binary features (signs) coded as 0 (absent) and 1 (present). Information from domain experts takes the form of the experts' estimate of the relative frequency of occurrence for each sign with respect to each class. The situation is far from ideal. First, no relationship between features is given. Second, the experts are asked to estimate frequencies for isolated features, out of the context of the other features relevant for the class. Precise numerical estimation of frequencies is a task that medical experts will seldom face in practice. Therefore, there are likely to be inaccurate estimates and divided opinions in devising the table of frequencies. Taking into account the shortcomings of these estimates, our aim is to develop a procedure which will shortlist a set of relevant features.

The most natural feature selection route beside the trivial choice of the individually best features is through simple sequential procedures, e.g., Sequential Forward Selection (SFS) and Sequential Backward Selection (SBS) [15]. In this paper we study the potential, limitations, stability and some implementation details of SFS for selecting features for non-traditional data, i.e., when the we only have probability estimates.

The rest of the paper is organised as follows. Section 2 explains the theory of feature selection for our problem and gives the amount of reduction on the classification error. Section 3 describes the feature selection procedure. Reliability of the selection procedure is discussed in Section 4 and a sensitivity analysis is proposed. Section 5 reports the result of our feature selection procedure for the differential diagnosis of Scrapie in sheep.

## 2 Theory

### 2.1 Feature selection

Feature selection is one of the oldest topics in pattern recognition and machine learning [15, 3, 9, 13]. Surveys on more recent state-of-the-art and comparisons between feature selection procedures can be found in [4, 2, 10, 1]. Classical feature selection procedures assume that we have a *traditional* data set. In our case we only have probability estimates and so are unable to pick combinations of features based on their interactions.

Let $X = \{x_1, \ldots, x_n\}$ be the set of binary features and $\omega_1, \ldots, \omega_c$, be the labels. We denote by $p_{j,k}$ the probability $p_{j,k} = P(x_j = 1|\omega_k)$. We will confine the study to two classes only. Surprisingly, even for this most elementary case with full knowledge of the distributions, selecting an optimal subset of features is not straightforward.

Since we do not have a traditional data set, the criterion for evaluating the feature subsets has to be calculated from the probabilities. In the literature, feature selection methods are grouped depending on the type of criterion function [4]

- *wrapper* methods, where the criterion is the classification accuracy of a classifier chosen for the problem at hand, when using only the features from the candidate subset;

- *filter* methods, where the criterion is another function, possibly accounting for the separability between the classes using the features from the candidate subset.

The wrapper approach has been found to be more successful than the filter approach [1]. This result is not surprising because the wrapper approach uses a direct measure of the performance characteristic of interest (classification accuracy) while the criteria used in the filter approach are only indirect measures, usually easier to compute than the accuracy. In this study we adopt the wrapper approach. With the two assumptions in place, the Naive Bayes classifier is the optimal classifier for non-traditional data and its accuracy will be the criterion for feature selection.

Sample size is an important issue in feature selection. Since we operate directly with probabilities, this issue is immaterial here.

### 2.2 Assumptions for non-traditional data

Our study is based on two assumptions which we cannot avoid given the information available for non-traditional data:

- The *independence* assumption. We assume that the features are independent given the class label,

i.e.,

$$P(\mathbf{x}|\omega_i) = \prod_{j=1}^{n} P(x_j|\omega_i), \quad i = 1, 2. \qquad (1)$$

This assumption does not often hold in practice. However, models based on a false independence assumption are reported to work well tolerating various degrees of dependencies [5]. We cannot avoid making this assumption because of our non-standard information set: there is no data from which we can estimate any dependencies between the features nor are experts willing/able to reliably estimate such dependencies.

- The *precision assumption*. We assume that the estimates of the probabilities given by the experts are accurate.

### 2.3 Sequential Forward Selection (SFS)

In *Sequential Forward Selection*, (SFS), we start with an empty set and add the best individual feature. To add feature $k + 1$, we check all possible feature sets of size $k+1$ which contain the $k$ features already selected and one feature from the remaining $n - k$. The feature that gives the best result is added to the set. The procedure stops when the desired number of features is reached or the error has been reduced to a specified target. The sequential procedures (SFS and the corresponding sequential backward selection (SBS)) have been found to be both simple and reasonably accurate. Variants thereof are deemed to be even more successful than the originals at the expense of a small increase of their computational complexity [1, 14]. For our problem, "best" subset means the subset with the minimum theoretical classification error.

We seek answers to the following questions

Question 1. *Is SFS monotone on the number of features?* That is, if $S$ is the selected feature set of size $k$, and $J(S)$ is the classification accuracy using the features from $S$, can we claim that

$$J(S \cup \{x_j\}) \geq J(S), \quad \forall x_j \in X \setminus S \ ? \qquad (2)$$

Question 2. *Is SFS optimal for our problem?* In other words, can we guarantee that the selected subset of features of size $d$ is the best subset of $d$ features selected from the original $n$ features under the problem assumptions?

Question 3. *Is the error reduction monotone?* Let $\Delta(x_j)$ be the error reduction when feature $x_j$ enters the set.[1] Can we claim that the largest error drop occurs at the first step, followed by decreasing $\Delta(x_j)$, $j = 2, 3, ...$? If this is the case, we can define a stopping criterion based on cost: we shall only add further features to the set if the error reduction outweighs the "cost" of measuring the feature; if the drop in the error becomes too small, then we can stop the selection arguing that further reductions does not justify more measurements.

Question 4. *How reliable are the results?* The probabilities that we use in the calculations are estimated by experts. How sensitive is the set of chosen features to changes in the probability estimates?

Questions 1 and 2 have been discussed in the literature of the 1960s and 1970s. Question 3 makes sense only if SFS is indeed monotone. Section 4 suggests a sensitivity analysis to answer Question 4.

### 2.4 The Naive Bayes classifier

Bayes classifier guarantees the minimum error [6]. The class label for an $\mathbf{x}$ is chosen to be the label corresponding to the largest posterior probability $P(\omega_i|\mathbf{x})$. Let $P(\omega_i)$ be the prior probability for class $\omega_i$, and $P(\mathbf{x}|\omega_i)$ be the class-conditional probability mass function for $\omega_i$. Consider $\mathbf{x}$ to be a binary vector of size $n$, i.e., $\mathbf{x} \in \{0,1\}^n$. The error for a particular $\mathbf{x}$ is $e(\mathbf{x}) = 1 - \max_i P(\omega_i|\mathbf{x})$. The total error ($\epsilon$) across $\{0,1\}^n$ is

$$\epsilon = \sum_{\mathbf{x} \in \{0,1\}^n} (1 - \max_i \{P(\omega_i|\mathbf{x})\}) P(\mathbf{x}) \quad (3)$$

where $P(\mathbf{x})$ is the unconditional probability mass function of $\mathbf{x}$. For two classes, $\omega_1$ and $\omega_2$, the expression reduces to

$$\epsilon = \sum_{\mathbf{x} \in \{0,1\}^n} e(\mathbf{x}) \quad (4)$$

where $e(\mathbf{x}) = \min\{P(\omega_1)P(\mathbf{x}|\omega_1), P(\omega_2)P(\mathbf{x}|\omega_2)\}$ is the error for the individual $\mathbf{x}$. Bayes classifier labels $\mathbf{x}$ in class $\omega_k$ if

$$P(\omega_k)P(\mathbf{x}|\omega_k) = \max_i P(\omega_i)P(\mathbf{x}|\omega_i). \quad (5)$$

Ties are resolved arbitrarily. For independent features, Bayes classifier becomes the so called "Naive Bayes", also guaranteeing the minimum error. Naive Bayes classifier labels $\mathbf{x}$ in class $\omega_k$ if

$$P(\omega_k) \prod_{j=1}^{n} P(x_j|\omega_k) = \max_i P(\omega_i) \prod_{j=1}^{n} P(x_j|\omega_i). \quad (6)$$

---
[1] Without loss of generality we can renumber the features in order of their entering the selected set using SFS.

Ties are resolved arbitrarily.

For example, consider a veterinary diagnostic problem with non-traditional data with $c = 2$ diseases and $n = 3$ signs. The following table contains the data $p_{ij}$, $i = 1, 2, 3$, $j = 1, 2$.

|       | $\omega_1$ | $\omega_2$ |
|-------|------------|------------|
| $x_1$ | 0.3        | 0.1        |
| $x_2$ | 0.4        | 0.6        |
| $x_3$ | 0.8        | 0.7        |

Suppose that an animal suspected of one of these diseases exhibits signs $x_1$ and $x_3$ but not sign $x_2$. Assume also that the two diseases are equiprobable, i.e., $P(\omega_1) = P(\omega_2) = 0.5$. The support for the two classes is calculated as follows

$$P(\omega_1|\mathbf{x}) \propto 0.5 \times 0.3 \times 0.6 \times 0.8 = 0.072 \quad (7)$$
$$P(\omega_2|\mathbf{x}) \propto 0.5 \times 0.1 \times 0.4 \times 0.7 = 0.014 \quad (8)$$

The Naive Bayes classifier will label $\mathbf{x}$ in class $\omega_1$.

The error of the Naive Bayes classifier for $\mathbf{x} \in \{0,1\}^n$ and $c = 2$ classes is

$$\begin{aligned} \epsilon &= \sum_{\mathbf{x} \in \{0,1\}^n} e(\mathbf{x}) \\ &= \sum_{\mathbf{x} \in \{0,1\}^n} \min \left\{ P(\omega_1) \prod_{j=1}^{n} P(x_j|\omega_1), \right. \\ &\qquad \left. P(\omega_2) \prod_{j=1}^{n} P(x_j|\omega_1) \right\}. \quad (9) \end{aligned}$$

### 2.5 Monotonicity of SFS

The so called "peaking phenomenon" or "Hughes paradox" occurs in feature selection with traditional data sets. It appears that there is an optimal number of features which give the smallest classification error: adding more features to the set raises the error. This finding is counterintuitive – the performance should not deteriorate when more information enters the model. The peaking phenomenon has been explained with the fact that only imperfect estimates of the probability distributions can be obtained with (finite) traditional data sets [3]. For non-traditional data, the following proposition holds.

> **Proposition 1**. *The theoretical error of the Bayes classifier does not increase when the feature set is augmented.*

Proposition 1 answers Question 1: SFS is monotone. The error cannot increase with adding a new feature to the set, regardless of whether or not this feature is an optimal choice. This proposition is regarded as a postulate in pattern recognition. Even though the proof is straightforward, it is seldom detailed in the literature. For completeness we show below the case of independent binary features and two classes.

Suppose we have selected $k$ features thereby creating a feature space $F$ with $2^k$ elements. Denote

$$\begin{aligned} a &= P(\omega_1)P(\mathbf{x}|\omega_1) \\ b &= P(\omega_2)P(\mathbf{x}|\omega_2) \\ c &= P(x_{k+1}=1|\omega_1) \\ d &= P(x_{k+1}=1|\omega_2) \end{aligned}$$

The error for $\mathbf{x}$ is $e(\mathbf{x}) = \min\{a,b\}$. By adding a new feature, $x_{k+1}$, we replace every $\mathbf{x} \in F$ by two new elements, $[\mathbf{x},0]^T$ and $[\mathbf{x},1]^T$, thereby doubling the number of elements of $F$. The error for $\mathbf{x}$ when feature $x_{k+1}$ is added splits into two:-

$$\begin{aligned} e([\mathbf{x}, x_{k+1}]) &= \min\{ac, bd\} \\ &+ \min\{a(1-c), b(1-d)\}. \end{aligned} \quad (10)$$

The reduction of the error for $\mathbf{x}$ is

$$\begin{aligned} \Delta(x_{k+1}) &= e(\mathbf{x}) - e([\mathbf{x}, x_{k+1}]) \quad (11) \\ &= \min\{a,b\} - (\min\{ac, bd\} \\ &+ \min\{a(1-c), b(1-d)\}) \quad (12) \end{aligned}$$

Using the representation $\min\{f,g\} = \frac{1}{2}(f+g-|f-g|)$, we arrive at the following expression

$$\begin{aligned} \Delta(x_{k+1}) &= \frac{1}{2}(|\underbrace{(a-b)}_{A} - \underbrace{(ac-bd)}_{B}| \\ &- |\underbrace{a-b}_{A}| + |\underbrace{ac-bd}_{B}|) \quad (13) \end{aligned}$$

Noticing that $|A| = |A - B + B| \leq |A - B| + |B|$, we conclude that

$$e(\mathbf{x}) - e([\mathbf{x}, x_{k+1}]) = \Delta(x_{k+1}) \geq 0. \quad (14)$$

∎

Note that the non-increasing of the error holds for every single $\mathbf{x} \in F$.

Equation (13) leads to various interesting results. Duin et al. [7] set up a threshold $\alpha(\mathbf{x}) = \frac{b}{a}$. If $\frac{c}{d}$ and $\frac{1-c}{1-d}$ both exceed $\alpha$ or are both smaller than $\alpha$ then there is no improvement. Duin et al. proceed to devise an ingenious diagram to illustrate this result. For a single $\mathbf{x}$, $\alpha(\mathbf{x})$ is a constant. The equations $\frac{c}{d} = \alpha$ and $\frac{1-c}{1-d} = \alpha$, plotted as lines in the plane $(c,d)$ define the region of no improvement. Figure 1 shows the no-improvement region (shaded) for $a = 0.3$ and $b = 0.7$. The new feature, $x_{k+1}$, is characterised by a pair $(c,d)$ and is a point on the diagram. (The dot at $(0.15, 0.75)$ and the triangle at $(0.90, 0.30)$ are two examples).

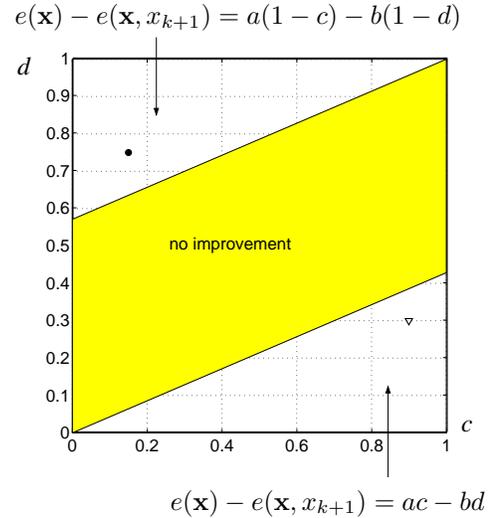

Figure 1: Regions of error reduction $\Delta(x_{k+1}, \mathbf{x})$ by introducing $x_{k+1}$ ($a = 0.3$, $b = 0.7$).

We can calculate from (13) the magnitude of the reduction of the error as shown in the figure. Hypothetically, the error can be reduced to 0 (maximum possible reduction), if $(c,d)$ is either $(0,1)$ or $(1,0)$. Thus the surface of the reduction is flat (zero) in the shaded region and has symmetrical plane "wings" which reach the maximum reduction (the outstanding error for this $\mathbf{x}$, $e(\mathbf{x})$) at the corners $(0,1)$ and $(1,0)$.

Feature $(k+1)$ will not contribute to the error reduction if and only if $(c,d)$ falls in the no-improvement regions for *all* elements of the current feature space $F$. The intersection of all no-improvement regions for binary features is found to be a parallelogram obtained from four lines: a pair of lines defined by the smallest $\alpha(\mathbf{x})$ greater than one and the pair of lines defined by the largest $\alpha(\mathbf{x})$ smaller than one ([7]). We note that the contribution for a particular $\mathbf{x}$ is not only a function of $(c,d)$ but also involves $a$ and $b$, which are specific for that $\mathbf{x}$. This shows that the contribution of a feature cannot be estimated in isolation even for the simple case of independent binary features and two classes. Thus a "league table" of important features can only be compiled cumulatively, i.e., subsequent features are estimated with respect to the features selected hitherto.

## 2.6 (Non-)Optimality of SFS

It has been proved in the early pattern recognition literature that even for two equiprobable classes and independent binary features, we cannot guarantee an optimal selection. A paper by Elashoff et al. [8] shows that the two independent individually best features might not be the best pair. Toussaint [16] extends the example to show that it is also possible that the best feature is not a even member of the best pair. The nonoptimality of SFS can be deduced from the Toussaint's result. SFS only adds features to the set and does not have a mechanism to remove features. Even if it did have such a mechanism, there will be no guarantee that the optimal set is found. It is known that the only type of procedures that guarantee the optimal solution are the exhaustive procedures (including branch-and-bound) [3].

### 2.7 (Non-)Monotonicity of the error reductions

The error of the selected set of features will decrease monotonically. Intuitively, we would expect that the largest drop of the error will be at the beginning of the procedure and smaller drops will be encountered toward the end. The example below shows that this is not true: a larger reduction can follow a smaller reduction.

Consider $P(\omega_1) = 0.3$ and $P(\omega_2) = 0.7$, as in the diagram in Figure 1. Let us label the two features depicted with markers as $x_1$ (the dot at $(0.15, 0.75)$) and $x_2$ (the triangle at $(0.90, 0.30)$). The error from using only the prior probabilities is $\min\{0.3, 0.7\} = 0.3$. The two individual errors are $e(x_1) = 0.22$ (reduction of 0.08 which can be calculated from the figure, $0.3(1 - 0.15) - 0.7(1 - 0.75) = 0.08$) and $e(x_2) = 0.24$ (reduction of 0.06). We select therefore $x_1$ first. By adding $x_2$, the error is reduced to 0.123 which gives a 0.097 drop, greater than the largest drop at the first stage.

The importance of each feature changes depending on which other feature enters the set. Suppose that we have two features $x_1$ and $x_2$ so that $x_1$ reduces the error for a given $\mathbf{x}$ (lies outside the shaded area in Figure 1) and $x_2$ does not reduce the error (lies inside the shaded area). It is possible that $x_2$ "escapes" from the no-improvement region after we add $x_1$ to the feature set. The following example demonstrates this. Let for the current $\mathbf{x}$, $a = 0.1$ and $b = 0.5$; thus $\alpha = \frac{b}{a} = 5$. Denote by $(c, d) = (0.4, 0.9)$ the probabilities that $x_1$ takes value 1 for classes $\omega_1$ and $\omega_2$, respectively, and by $(e, f) = (0.1, 0.8)$, the corresponding probabilities for $x_2$. By checking the conditions of Duin et al. [7], we find out that $(c, d)$ does reduce the error ($\frac{c}{d} < 5$ and $\frac{1-c}{1-d} > 5$) while $(e, f)$ does not ($\frac{e}{f} < 5$ and $\frac{1-e}{1-f} < 5$). Taking $x_1$ in the set, two new elements of the feature space $F$ are generated from $\mathbf{x}$. For $(\mathbf{x}, x_1 = 1)$, the new $a$ and $b$ are $a' = ac$ and $b' = bd$. The new $\alpha$ is therefore $\alpha' = 11.25 > \alpha$. Then the new no-improvement region expands and contains all the points (features) in the old no-improvement region. For the second new point, $(\mathbf{x}, x_1 = 0)$, the new $a$ and $b$ are $a'' = a(1 - c)$ and $b'' = b(1 - d)$. The new $\alpha$ is therefore $\alpha'' = \frac{5}{6} < \alpha$. Now feature $x_2$ is outside the no-improvement region because $\frac{e}{f} < \frac{5}{6}$ and $\frac{1-e}{1-f} > \frac{5}{6}$.

It is even more curious that even when no reduction is indicated at some stage of SFS by any of the remaining features, we can enter a "redundant" feature (no error reduction) which will enable further reduction by other features.

The non-monotonicity of reduction means that we cannot use the size of the reduction as a simple stopping criterion. The outstanding error might vanish in the next step or might be reduced in small quantities over a long selection process. Also, if at some stage there is no feature that improves on the error but we have not reached the desired number of features, we should continue with SFS by selecting a feature among the non-relevant ones. The best candidates would be the features close to the boundaries of the non-improvement region.

## 3 Implementation issues

Even though SFS is non-optimal, there is a strong argument in the literature in favour of sequential procedures and their variants [1]. For problems such as ours, which is heavily assumption-bound, we chose SFS because of its simplicity. Our experiments showed that the predicted reduction on the error using the features selected through SFS was quite substantial. Further reduction by using a more complicated procedure might not be justified because the true reduction will depend mostly on the validity of the assumptions.

This section gives the technical details of the SFS implementation. We use the standard procedure for adding one feature at a time. Below we explain the way to calculate the criterion value.

The fastest way to calculate the error is to maintain a list with the elements of the current feature space with the corresponding $a$ and $b$ values (see previous section). The list starts with just one element containing the prior probabilities, i.e. $(a, b) = (P(\omega_1), P(\omega_2))$. To check a new feature $x_{k+1}$, the list is expanded by creating two elements in the place of each single element of the feature space. The new elements have parameters $(ac, bd)$ and $(a(1-c), b(1-d))$. For example, suppose we start with $L = (0.3, 0.7)$. When we add $x_1$ with $c = 0.15, d = 0.75$, the single point in $L$ is replaced by two new points, corresponding to $x_1 = 1$ and $x_1 = 0$. The list is modified to

$$L = \begin{bmatrix} 0.3 \times 0.15 = 0.045 \\ 0.7 \times 0.75 = 0.525 \\ 0.3 \times 0.85 = 0.255 \\ 0.7 \times 0.25 = 0.175 \end{bmatrix} \quad (15)$$

The error is calculated using (10), in this example $e(x_1) = 0.045 + 0.175 = 0.22$. This implementation is fast but space-consuming as the list contains $2^k$ elements and we need to store two values for each. Below is a MATLAB implementation of the procedure that updates the list $L$ and calculates the error for introducing a new feature.

```
function [e,e_fp,e_fn,L_new]=update_L(L,p,q);
[m,n]=size(L);
L_new=[L.*repmat([p q],m,1);
L.*repmat([(1-p),(1-q)],m,1)];
[L_new_min,index]=min(L_new');
e=sum(L_new_min);
e_fp=sum(L_new_min(index==2));
e_fn=e-e_fp;
```

In the code p is $P(x_{k+1} = 1|\omega_1)$ and q is $P(x_{k+1} = 1|\omega_2)$. Suppose that $\omega_1$ is the class label corresponding to the disease of interest. Along with the error e, the program outputs its two components: the false positives $\mathtt{e\_fp} = P(\text{Positive test} \cap \text{Non-disease})$ and the false negatives $\mathtt{e\_fn} = P(\text{Negative test} \cap \text{Disease})$. The updated list L_new is twice longer than the input list L. The function is called to check the error for every available feature as $x_{k+1}$. The list which corresponds to the smallest error is retained and the respective feature is added to the set.

*Sensitivity* and *specificity* of the selected feature set can be calculated from the two components of the error

$$\text{Sensitivity} = \frac{\text{Positive test}}{\text{All true positive}} = \frac{P(\omega_1) - e_{fn}}{P(\omega_1)} \quad (16)$$

$$\text{Specificity} = \frac{\text{Negative test}}{\text{All true negative}} = \frac{P(\omega_2) - e_{fp}}{P(\omega_2)}. \quad (17)$$

Evaluation of the criterion can be implemented in a recursive way. This approach will require less memory/space but will take substantially longer than the linear implementation.

## 4 Stability of SFS

The adequacy of our feature selection will depend mainly on the validity of the assumptions. The robustness of the proposed technology with respect to violation of the independence assumption cannot be easily assessed. The reason is that modelling dependencies that might be far away from the real dependencies between the features will not give us much insight into the real-life performance of SFS. However, we can test SFS on perturbed values of the frequencies.

For each frequency separately we consider truncated normal distributions with mean equal to the expert estimate and standard deviations $\sigma = \{0.1, 0.2, 0.3\}$. The idea is to run SFS, say, 1000 times on the perturbed frequencies and to find out how similar the selected feature sets are.

One way to measure the result from this experiment is to calculate the features' ranks. Consider a single SFS run for $d = 10$ features out of $n >> d$ features. A feature gets rank 10 if it is selected first, rank 9 if it selected second, etc., rank 1 if it selected in the 10th place and rank 0 if it is not selected. After 1000 SFS runs on frequencies generated from the truncated normal distributions, the ranks for each feature are summed up. If a feature has been selected first in all 1000 experiments, then its sum of ranks (total rank) would be 10,000. If its total rank is below 1000, this would mean that the feature has not been selected in all 1000 sets. (Note that rank above 1000 does not mean that the feature has been selected in all 1000 sets.) The total rank of a feature would be a rough indication of its discrimination value. A high match in the feature lists (sorted by rank) coming from the distributions with the three values of $\sigma$ will signify a robust feature selection procedure.

Even more interesting could be the mismatches. If small perturbations bring new features in the set of "important" features, then these new features might also be worth a second look. On the other hand, if due to small perturbations some features drop off the list of important features, this might mean that these features are not too reliable.

## 5 Feature selection for diagnosing Scrapie in sheep

Scrapie is an endemic transmissible spongiform encephalopathy of sheep in Great Britain although not all flocks are affected. Scrapie was made a notifiable disease in 1993 [11]. Compensation and compulsory slaughter of scrapie suspects were introduced in 1998 [11].

There is currently no ante-mortem test for scrapie currently available for routine field use. The differential diagnosis of a scrapie suspect could be assisted by the provision of a list of clinical signs that have the greatest discriminatory power to categorise suspects into scrapie and non-scrapie categories. This may reduce the number of false positives suspects and, more importantly reduce the number of false negatives suspects. Information regarding the sign frequencies for these differential diagnoses in the literature is sparse and inconsistent. Expert opinion has been used to generate disease sign frequencies with useful results in veterinary medicine [12]. The opinion of 3 specialists in sheep health and production were used in the study. Each expert provided clinical sign frequencies for the signs encountered within each of the 62 differential diagnoses and scrapie. The mean value of the 3 experts was used in the study.

Table 1: The first 15 signs selected by SFS and the cumulative error, sensitivity and specificity (in %).

| # | Feature | Error | Sens | Spec |
|---|---|---|---|---|
| 1. | Hyperaesthesia | 9.87 | 86.7 | 93.6 |
| 2. | Weight loss. | 7.47 | 86.7 | 98.4 |
| 3. | Pruritus | 2.93 | 97.3 | 96.8 |
| 21. | Increased respiratory rate | 2.63 | 97.3 | 97.4 |
| 4. | Abnormal behaviour | 2.21 | 99.3 | 96.3 |
| 5. | Underweight | 1.37 | 98.3 | 98.9 |
| 9. | Tremor | 1.18 | 99.3 | 98.3 |
| 22. | Sudden death | 1.02 | 99.3 | 98.6 |
| 6. | Dysmetria | 0.92 | 99.3 | 98.9 |
| 7. | Ataxia | 0.67 | 99.3 | 99.3 |
| 8. | Grinding teeth | 0.55 | 99.5 | 99.4 |
| 10. | Trembling | 0.45 | 99.6 | 99.5 |
| 11. | Alopecia | 0.36 | 99.6 | 99.7 |
| 12. | Seizures or syncope | 0.32 | 99.7 | 99.7 |
| 13. | Rumen hypomotility | 0.29 | 99.7 | 99.7 |

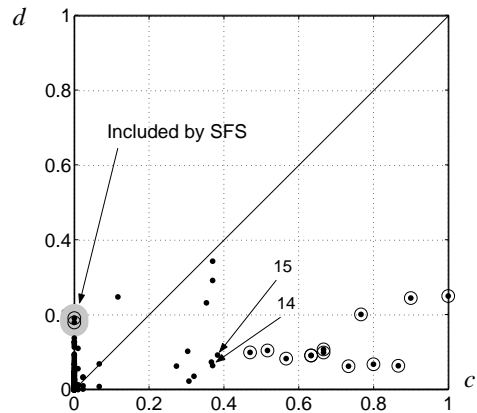

Figure 2: $(c, d)$-points corresponding to all 285 features for the Scrapie problem. The 15 features selected through SFS are encircled.

The set of frequencies provided by the experts consists of 285 binary features (signs) with regard to Scrapie and 62 alternative diseases. We averaged the frequencies across the alternative diseases to obtain $p_{j,1}$ and $p_{j,2}$, $j = 1, \ldots, 285$, assuming equal prevalences. Using this information we applied SFS for selecting the best 15 features out of the original 285 features. The results are given in Table 1. The features are numbered with respect to their individual importance. Feature #1 has the greatest absolute difference between $P(x_j = 1|\text{Scrapie})$ and $P(x_j = 1|\text{Non-Scrapie})$ among all 285 features. The cumulative error, sensitivity and specificity [all in %] are shown.

The features that have been selected are shown in the order they entered the set. Note that the selected set of 15 contains the individually best 13, and in almost the correct order of entering. Choosing the individually best 15 features gives an error of 0.33%, which is slightly higher than the error of the set selected through SFS. The difference is actually too small to merit a claim that SFS has selected a better subset.

Next we ran the sensitivity analysis as explained in the previous section. Although the order of selecting the features was slightly different, the lists for the three values for $\sigma$ were highly similar. The top 10 ranked features for all three lists were the top ten individual features in a slightly different order.

A comparison between SFS on the one hand and SBS, branch-and-bound and exhaustive search on the other hand, using the true error as the selection criterion, is computationally prohibitive for the current problem with $n = 285$ features. The fact that SFS picked most of the individually best features is not surprising. The best features are those whose corresponding points in Figure 1 would be the closest to either top left or bottom right corner. These points are less likely to fall in the region of no improvement. The points for the 285 features for the Scrapie problem are depicted in Figure 2. The 15 features selected by SFS are encircled. The independence assumption guarantees that the changes in the no-improvement region will be independent on the features which are not yet selected. Therefore features which are away from the diagonal line have a good chance to be far enough from the no-improvement region.[2]

The effect of SFS can be seen in choosing the two features on the left edge of the graph. These features became important in relation to the other features already selected. On the other hand, features ranked 14 and 15 were dismissed by SFS. It is possible that the advantages of SFS will appear when more features are selected.

## 6 Conclusions

In differential diagnosis of a new or a rare disease the available information is often in the form of human es-timates of the frequencies of occurrence of signs and symptoms. We study the SFS feature selection procedure for the so-called "non-traditional" data where the information is in the form of expert estimates of the conditional probabilities of occurrence of the features.

---

[2]Notice that only the region of no-improvement will be different; the point position is the same, only the relative importance of the points changes with the changes of the region.

**The independence assumption.** The Naive Bayes model is adopted for the feature selection assuming independence between features. Since there is no (traditional) data set, assuming any specific dependency without a verification would be only a speculation. Domingos and Pazzani [5] argue that deviations from the assumption of independence are not too harmful to the performance of the Naive Bayes model.

**The SFS procedure.** SFS is generally perceived to be simple, robust and reasonably accurate. We suggest that possible inadequacies of the selected set of features would not be due to the simplicity of SFS but due to the invalid assumptions of the underlying model. SFS was discussed in Section 2 with respect to its optimality and monotonicity. Although SFS does not guarantee the optimal feature subset, it is a good practical solution. In the experiments in Section 5, SFS drove the classification error to less than half percent with just 11 out of the original 285 features. Some tips about the technical implementation of SFS for independent binary features and non-traditional data sets were given in Section 3.

**The expert estimates.** Our proposed model will only be as valid as the underlying estimates of the probabilities are. If the estimates are far from the true probability values, the model might not perform too well. The sensitivity analysis proposed in section 4 aims at finding out the robustness of SFS in selecting a feature subset when the estimates of the frequencies vary according to a truncated normal distribution within the interval [0,1]. The results given in Section 5 show that SFS is indeed robust. In all the experiments the selected sets consisted primarily of the features that were found to be individually relevant. This may be partly due to the independence assumption, although the theory shows that such a result is not guaranteed.

Another approach to verifying these probabilities would be to use meta analysis on published results.

The most important question still to be answered is whether the selected feature set will work with real data. The problem is that such a data set has to be collected first. It may be worthwhile expanding the set of relevant features to a practically reasonable number, including the cost of measuring, so that the collection of data is possible on a regular basis and no important feature is missed.